\documentclass[letterpaper, 10 pt, conference]{ieeeconf}

\IEEEoverridecommandlockouts                              

\usepackage{graphicx}
\usepackage{amsmath}
\usepackage{amssymb}
\usepackage{caption}
\usepackage{subcaption}
\usepackage{hyperref}
\usepackage[table]{xcolor}
\usepackage{draftwatermark}
\usepackage[final,markup=bfit, deletedmarkup=sout, authormarkup=brackets]{changes}

\usepackage{array,multirow,graphicx}

\graphicspath{ {figs/} }
\usepackage{tikz}

\begin{document}
\SetWatermarkAngle{0}
\SetWatermarkColor{black}
\SetWatermarkLightness{0.5}
\SetWatermarkFontSize{10pt}
\SetWatermarkVerCenter{20pt}
\SetWatermarkText{\parbox{30cm}{%
\centering This is the final version of the manuscript accepted for publication in\\
\centering `2021 Robotics and Automation (ICRA), IEEE International Conference on'. (C) IEEE.}}

\title{\LARGE \bf 3D Collision-Force-Map for Safe Human-Robot Collaboration}

\author{Petr~Svarny,
        Jakub~Rozlivek,
        Lukas~Rustler,
        and~Matej~Hoffmann
\thanks{All authors are with Department of Cybernetics, Faculty of Electrical Engineering, Czech Technical University in Prague; \texttt{\{petr.svarny, rozlijak, rustlluk, matej.hoffmann\}@fel.cvut.cz}. This work was supported by the Czech Science Foundation (GA CR), project EXPRO (no. 20-24186X). P.S., J.R., and L.R. were additionally supported by the Czech Technical University in Prague, grant no. SGS20/128/OHK3/2T/13. 
The authors would like to thank Minh Thao Nguyenova, Martin Sramek, and Petr Posik for their valuable suggestions and Abdel-Nasser Sharkawy for assistance with the effective mass analysis.}
}

\author{Petr~Svarny,
        Jakub~Rozlivek,
        Lukas~Rustler,
        and~Matej~Hoffmann%
\thanks{
This work was supported by the Czech Science Foundation (GA CR), project EXPRO (no. 20-24186X). P.S., J.R., and L.R. were additionally supported by the Czech Technical University in Prague, grant No. SGS20/128/OHK3/2T/13. 
The authors would like to thank Minh Thao Nguyenova, Martin Sramek, and Petr Posik for their valuable suggestions and Abdel-Nasser Sharkawy for assistance with the effective mass analysis.} 
\thanks{All authors are with Department of Cybernetics, Faculty of Electrical Engineering, Czech Technical University in Prague.
        {\tt\footnotesize \{petr.svarny, rozlijak, rustlluk, matej.hoffmann\}@fel.cvut.cz}}%
\thanks{Digital Object Identifier (DOI): see top of this page.}
}
%
%

\markboth{IEEE Robotics and Automation Letters. Preprint Version. Accepted Month, 2021}
{Svarny \MakeLowercase{\textit{et al.}}: 3D Collision-Force-Map} 
%


\maketitle


\begin{abstract}
    The need to guarantee safety of collaborative robots limits their performance, in particular, their speed and hence cycle time. The standard ISO/TS 15066 defines the Power and Force Limiting operation mode and prescribes force thresholds that a moving robot is allowed to exert on human body parts during impact, along with a simple formula to obtain maximum allowed speed of the robot in the whole workspace. In this work, we measure the forces exerted by two collaborative manipulators (UR10e and KUKA LBR iiwa) moving downward against an impact measuring device. First, we empirically show that the impact forces can vary by more than 100 percent within the robot workspace. The forces are negatively correlated with the distance from the robot base and the height in the workspace. Second, we present a data-driven model, 3D Collision-Force-Map, predicting impact forces from distance, height, and velocity and demonstrate that it can be trained on a limited number of data points. Third, we analyze the force evolution upon impact and find that clamping never occurs for the UR10e. We show that formulas relating robot mass, velocity, and impact forces from ISO/TS 15066 are insufficient---leading both to significant underestimation and overestimation and thus to unnecessarily long cycle times or even dangerous applications. We propose an empirical method that can be deployed to quickly determine the optimal speed and position where a task can be safely performed with maximum efficiency.
\end{abstract}

\IEEEpeerreviewmaketitle

\section{Introduction}
\label{sec:intro}
\textit{Physical Human-Robot Interaction} (pHRI) or \textit{Human-Robot Collaboration} (HRC) (e.g., \cite{Haddadin2016}) is a dynamically growing research field. At the same time, in industry, the expectations associated with collaborative robots (or co-bots; robots designed for direct interaction with humans~\cite{ISO8373}) are high, but their uptake has been somewhat held up by their performance limitations derived from strict safety constraints.
Various safety standards, especially ISO 10218~\cite{ISO10218} and ISO/TS 15066~\cite{ISO/TS15066} (TS 15066 for short), formulate these safety demands. These standards currently list four modes of collaboration. While all HRC is ``continuous, purposeful interaction associated with potential or accidental physical events''~\cite{Vicentini2020a}, only the Power and Force Limiting (PFL) mode permits physical contact between the robot and the human when the robot is still autonomously moving, provided that the impact force, pressure, and energy stay within prescribed limits \cite{ISO/TS15066}. Nevertheless, the force thresholds derived from the TS 15066 for the PFL mode enforce low operational velocities, especially if there is a risk of clamping (see for example \cite{Mansfeld2018} or \cite{Lucci2020}). These constraints motivate current investigations geared to overcoming the limitations. 

\begin{figure}[h]
\centering
\begin{subfigure}{0.2\textwidth}
\includegraphics[height=3cm]{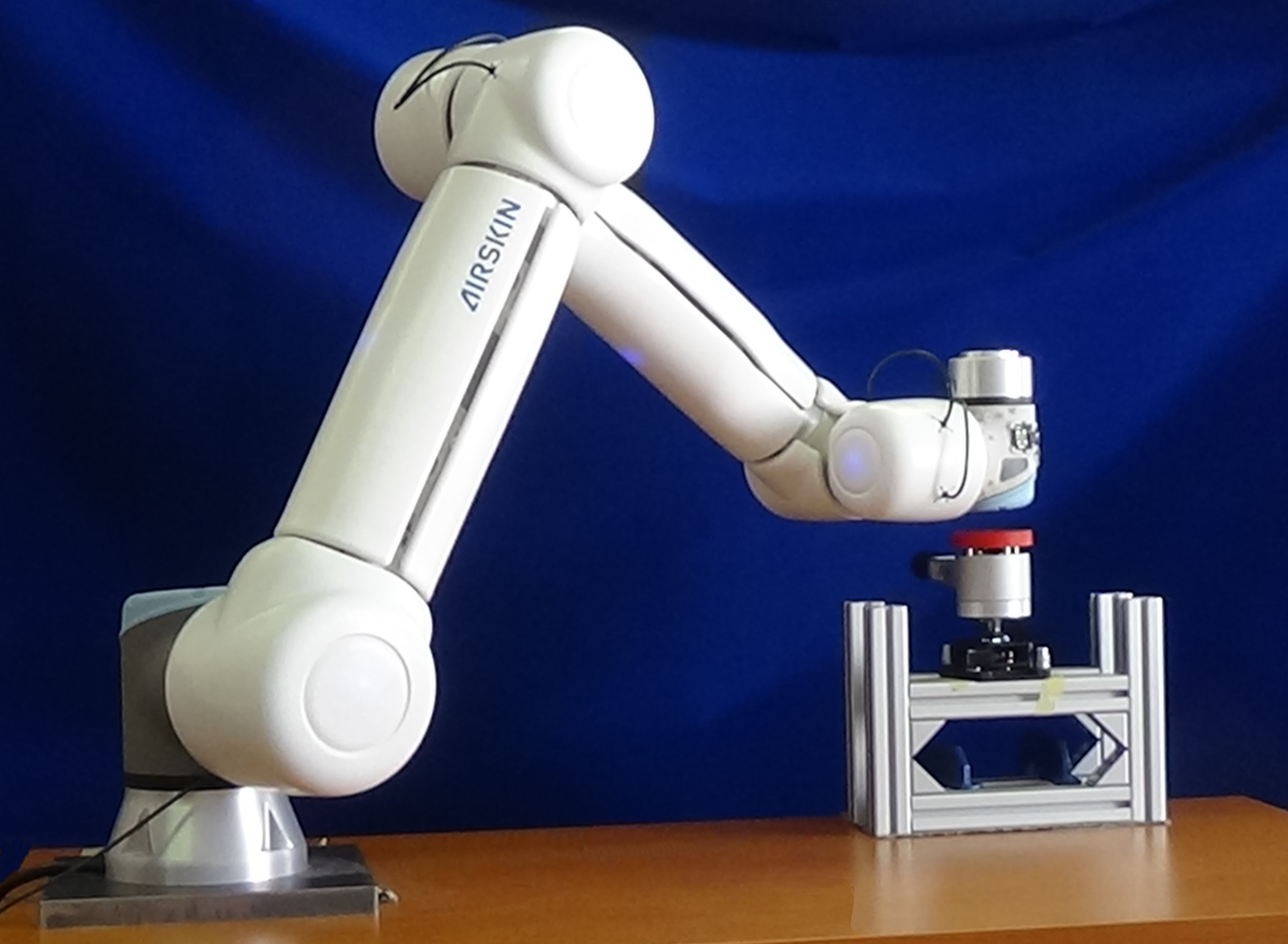} 
\caption{{\small UR10e.}}
\label{fig:ur_setup}
\end{subfigure}
\hfill
\begin{subfigure}{0.18\textwidth}
\includegraphics[height=3cm]{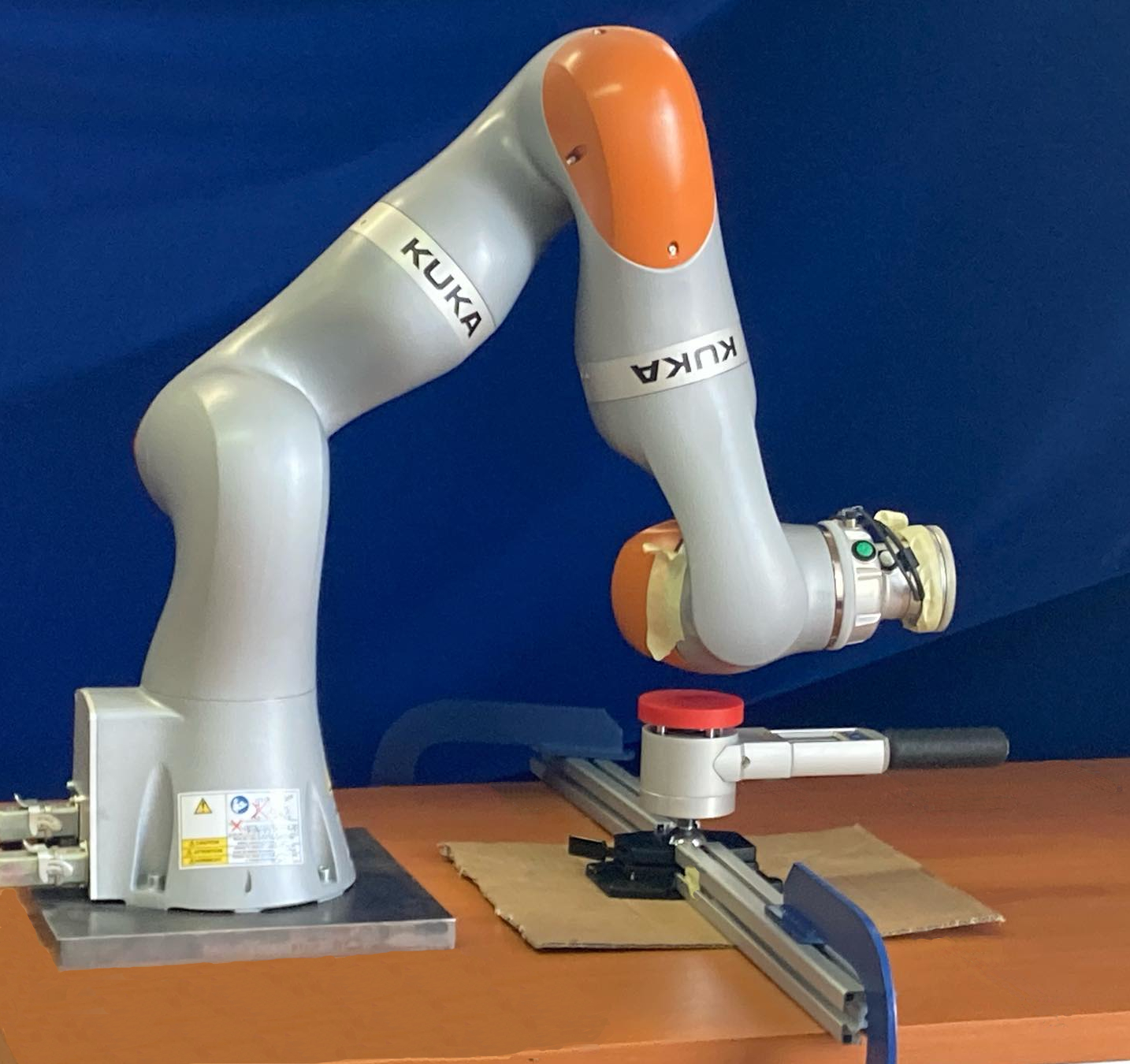}
\caption{{\small KUKA LBR iiwa.}}
\label{fig:kuka_setup}
\end{subfigure}
\vspace{-5pt}
\caption{{\small Setup -- robots and impact measuring device.}}
\label{fig:setup}
\vspace{-0.2cm}
\end{figure}

TS 15066 prescribes limits based on pain thresholds from studies like \cite{Suita1995}. These limits are, however, subject to a heated debate, e.g. \cite{Mansfeld2018}, \cite{Han2018}, \cite{Park2019}.  

A detailed treatment of safety aspects of human-robot collisions is presented in \cite{Haddadin2015, Haddadin2017}. Contact modeling \textit{per se} is notoriously difficult; in HRC, it is even more challenging as many parameters (mass and its distribution in colliding bodies, behavior of robot controller upon impact, etc.) are not known. Kovincic et al. \cite{Kovincic2019,Kovincic2020} suggest using collected impact data to model the impact forces using machine learning approaches because robot reaction mechanisms play a significant role in the resulting forces and are ``not known or can not be identified''~\cite{Kovincic2019}. Schlotzhauer et al.~\cite{Schlotzhauer2019} introduce a 2D Collision-Force-Map (2D CFM) and approximate the impact forces of UR10 and UR10e robots in a pick and place task with a second degree polynomial.

Finally, robot performance can be further boosted if PFL is not treated in isolation. For example, combinations with the Speed and Separation Monitoring collaboration mode---where robot needs to come to stop before contact---can be implemented (e.g., \cite{Svarny2019}). This can take the form of optimal velocity scaling \cite{Lucci2020}, velocity scaling based on an impact force model \cite{Shin2018}, the use of control barrier functions \cite{Ferraguti2020}, or by predicting the exerted force based on motor currents \cite{Aivaliotis2019}.

This work focuses on the PFL collaborative mode. We measure the forces exerted by two collaborative manipulators (UR10e and KUKA LBR iiwa) on an impact measuring device in different positions in the robot workspace and with various velocities (Fig.~\ref{fig:setup}). Our approach is similar to 2D CFM~\cite{Schlotzhauer2019} in that we use empirical measurements and fit a function relating robot position and speed to the impact force. Newly, we establish the importance of the height, leading to a 3D Collision-Force-Map (3D CFM).

\textbf{Our contributions are the following}: (i) a 3D collision-force-map is created, considering the velocity, distance from robot base, and, newly, the height in the workspace; (ii) a simple data-driven model using only few samples is presented and validated; (iii) behavior of the two manipulators upon impact is analyzed, drawing important implications for their deployment in collaborative applications. 

This article is structured as follows. The related theory is in Section~\ref{sec:CFM_effective_mass}. The Experimental setup section is followed by Experiments and Results (\ref{sec:results}). We close with Conclusion, Discussion, and Future work.

An accompanying video illustrating the experiments is available at {\footnotesize\url{https://youtu.be/4eHsbe4EuHU}}. The dataset with the data collected is at \cite{public-dataset}.

\section{Collision force maps and effective mass}
\label{sec:CFM_effective_mass}

\subsection{Power and Force Limiting}
\label{subsec:PFL}

\begin{figure}[bt]
	\centering
	\includegraphics[width=0.3\textwidth]{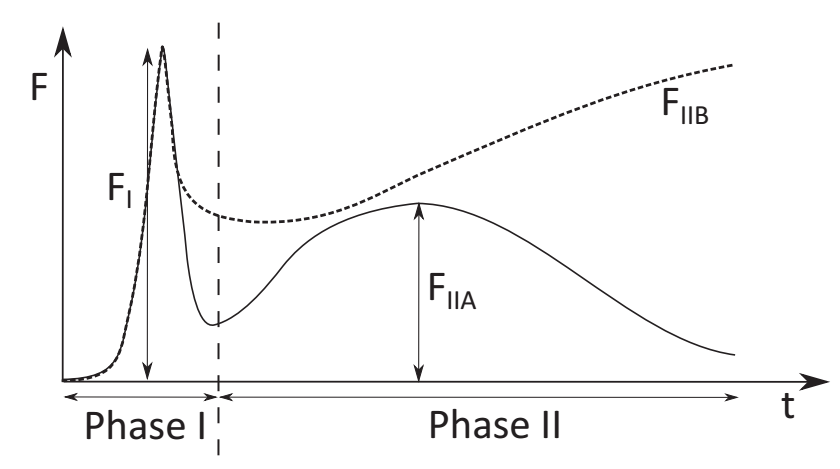}
    \vspace*{-5pt}
	\caption{{\small Collision phases from \cite{Haddadin2015}. Phase I, the initial dynamic impact with the force $\mathrm{F_{I}}$, and Phase II,  either a diminishing force profile $\mathrm{F_{IIA}}$ in the case of no clamping or a non-diminishing force profile $\mathrm{F_{IIB}}$ if there is clamping.}}
	\label{fig:phases}
    \vspace{-0.6cm}
\end{figure}

A human-robot collision can be decomposed into two phases (see \cite{Haddadin2015}). An initial dynamic impact in Phase I is followed by the Phase II force profile that depends on the clamping nature of the incident (see Fig. \ref{fig:phases}).
There are at least three possible scenarios (for details see \cite{Vicentini2020a}):
\begin{itemize}
    \item unconstrained dynamic impact (no force in Phase II),
    \item constrained dynamic impact without clamping (diminishing force in Phase II)
    \item constrained dynamic impact with clamping (force is not diminishing in Phase II)
\end{itemize}

TS 15066 \cite{ISO/TS15066} does not make a distinction between the impact phases and merely distinguishes between two scenarios, a \textit{transient} contact, i.e. dynamic impact that is unconstrained or is not followed by clamping, and \textit{quasi-static} contact, i.e. dynamic impact followed by clamping.
The equation A.6 from TS 15066 relating velocity ($v$) and (maximum) impact force ($F_{\mathrm{max}}$) is:
\begin{equation}
    v \leq  \frac{F_{\mathrm{max}}}{\sqrt{k}}\sqrt{m_{R}^{-1} + m_{H}^{-1}},
    \label{eq:v_pfl}
\end{equation}
with $m_R$ the effective robot mass, $m_H$ the human body part mass, $k$ the spring constant for the human body part and $F_{\mathrm{max}}$ the maximum impact force permitted for the given body region. As pointed out in \cite{Mansfeld2018}, this is a simplified contact model with a single spring constant for the human body.

However, the risks cannot be evaluated solely based on the robot---it is necessary to take into account the application as a whole.
We assume a mock pick and place scenario with a risk of a constrained dynamic impact on the human hand as in \cite{Schlotzhauer2019}. 
Contact may occur as the robot is descending towards the table, possibly clamping the hand of the operator. In practice, a risk analysis according to \cite{ISO12100} will be required.

If we investigate constrained dynamic impacts, even without clamping, we can approximate $m_H^{-1}\approx 0$ as in \cite{Lucci2020}\footnote{The impacted body part is constrained and thus immovable. Its weight in the PFL two-body spring model can be considered therefore as significantly larger than the other body's, and hence approximated as infinite.}. This approximation allows us also to simplify the situation by investigating the relative velocity as simply the robot velocity with the human hand being still. The other variables are therefore set based on \cite{ISO/TS15066} as $F_{\mathrm{max}}=\ $140 N and $k=\ $75000 N/m. The moving masses of the UR and KUKA robot are approximately 30 kg and 20 kg respectively. Using the approximation from \cite{ISO/TS15066} that the effective robot mass $m_R$ is $M/2 + m_L$ (half of the total mass of the moving parts of the robot, plus the effective payload $m_L$, which is zero in our case), together with Eq.~\ref{eq:v_pfl}, would give permissible velocity up to 0.13 m/s for the UR robot and 0.16 m/s for the KUKA robot in case of clamping. If there is no clamping, the permissible force becomes 280 N and thus also the velocities are higher, namely 0.26 m/s for the UR and 0.32 m/s for the KUKA due to the weight difference between the robots.

\subsection{Collision-Force-Map -- 2D and 3D}
\label{subsec:CFM_2D_3D}
The assumptions and approximations made in \cite{ISO/TS15066} are too coarse and do not match empirical impact measurements.
Schlotzhauer et al.~\cite{Schlotzhauer2019} proposed a 2D Collision-Force-Map---a data-driven linear model to predict the impact force as a function of the distance from the z-axis of the robot base frame ($d$) and velocity ($v$). The model is a second degree polynomial of the form:
\begin{equation}
    \mathrm{ln}(F) =  \beta_{0} + \beta_{1}\cdot v + \beta_{2} \cdot d + \beta_{3} \cdot d^2
    \label{eq:linear_model}
\end{equation}
The parameters are robot-, software-, and application-specific and should be found from a large number of measurements.

\begin{figure}[tb]
 	\centering
 	\includegraphics[width=0.45\textwidth]{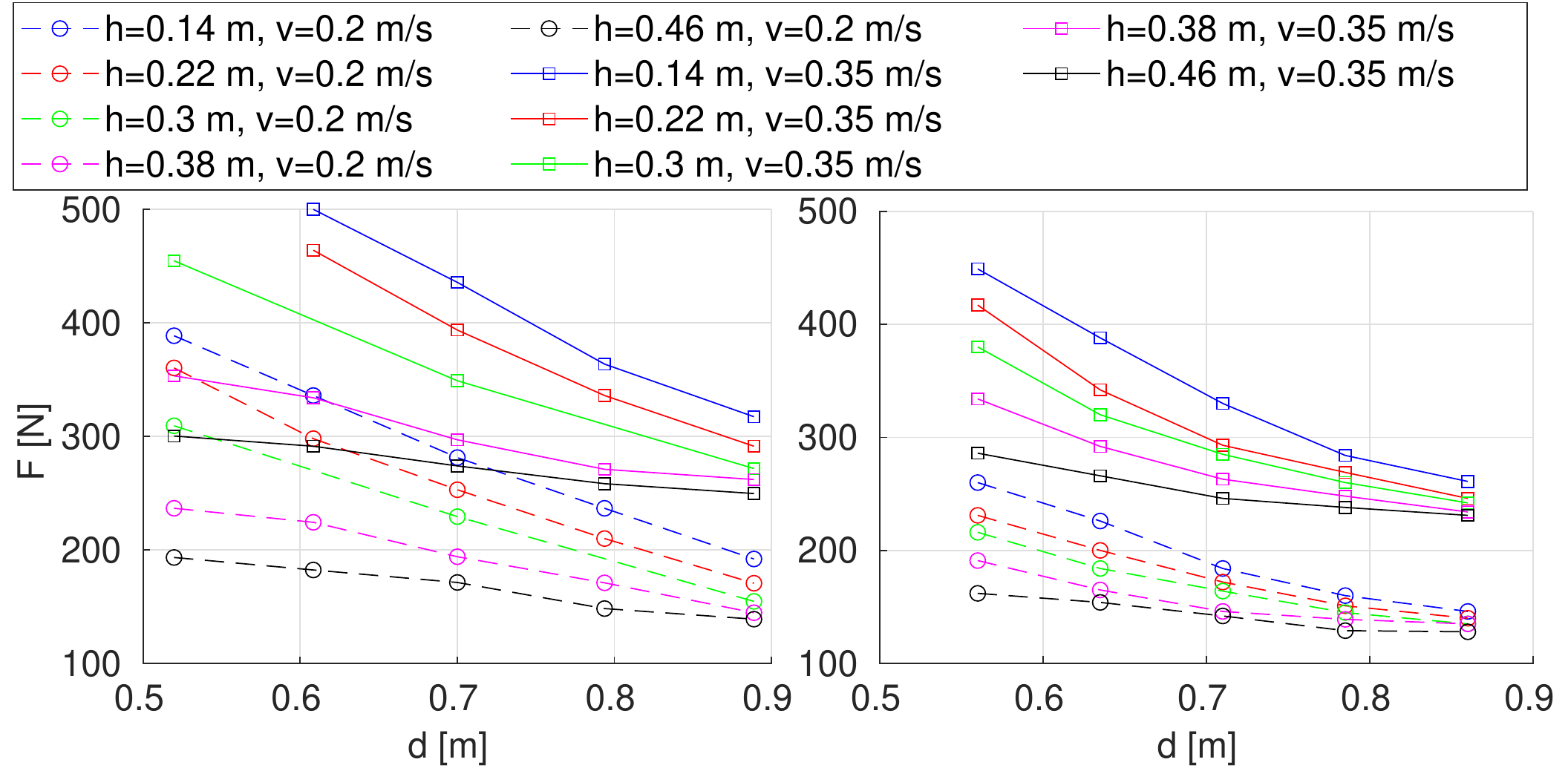}
 	\vspace*{-5pt}
 	\caption{{\small Impact forces for different distances, heights, and velocities of EE. (Left) UR10e. (Right) KUKA LBR iiwa 7R800 with 10 Nm external torque limit. 
 	}}
 	\label{fig:force_distance_height}
 	\vspace{-0.3cm}
\end{figure}

In this work, we add the height in the workspace \added{($h$)} as an important additional dimension that affects the force exerted on impact. Euclidean distance in 3D between the end effector~(EE) and the robot base would be a candidate representation, leading to a different 2D Collision-Force-Map. However, our empirical measurements---see Fig.~\ref{fig:force_distance_height}---reveal a more complicated relationship between $d$, $h$, and $v$. For the UR10e robot (Left), the dependence of force on distance has a different profile for $h \geq $ 0.38$\ \mathrm{m}$ than for lower heights. This is true for two different speeds. For the KUKA (Right), the contribution of height to predicting the impact forces goes down with the distance from the base.   

\subsection{Effective mass as function of d and h}
\label{subsec:effective_mass}

We sought a theoretical rationale for the observations above. While TS 15066 considers the ``effective robot mass'' statically, Khatib~\cite{khatib1995inertial} introduced the robot effective mass as a dynamic property depending on the robot's configuration and the impact direction. This has been later adopted by many others, sometimes also called reflected mass (e.g., \cite{Haddadin2016,Mansfeld2018,Lucci2020,Haddadin2015,Lee2013}).      
The effective mass of a manipulator in a given direction $\boldsymbol{u}$ can be modeled using the formula \cite{Lee2013}:
\begin{equation}
    m_{\boldsymbol{u}}^{-1} = \boldsymbol{u}^T [J(\boldsymbol{q}) M^{-1}(\boldsymbol{q}) J^{T}(\boldsymbol{q})] \boldsymbol{u},
    \label{eq:eff_mass}
\end{equation}
where $\boldsymbol{q}$ are the joint angles of a given position, $M(\boldsymbol{q})$ and $J(\boldsymbol{q})$ are the
Inertia matrix and the Jacobian matrix of the manipulator, respectively (see, e.g.,  \cite[Ch. 3 and Ch. 7]{Siciliano2010}).

\begin{figure}[tb]
\centering
\begin{subfigure}{0.225\textwidth}
\centering
\includegraphics[width=\textwidth]{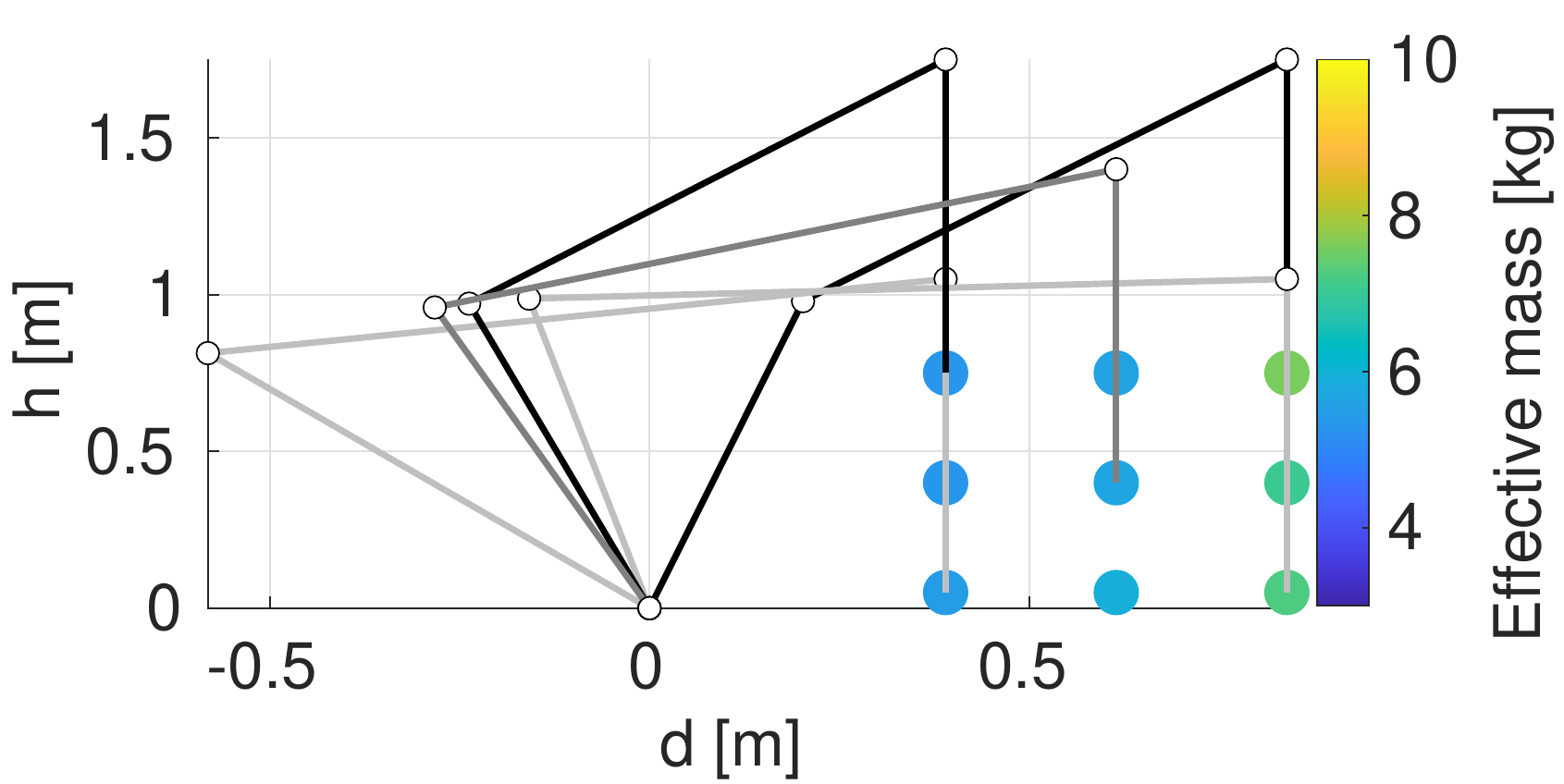} 
\vspace{-0.6cm}
\caption{{\small $m_R$ at 9 inspection points.}}
\label{fig:3dof_3by3}
\end{subfigure}
~
\begin{subfigure}{0.24\textwidth}
\centering
\includegraphics[width=\textwidth]{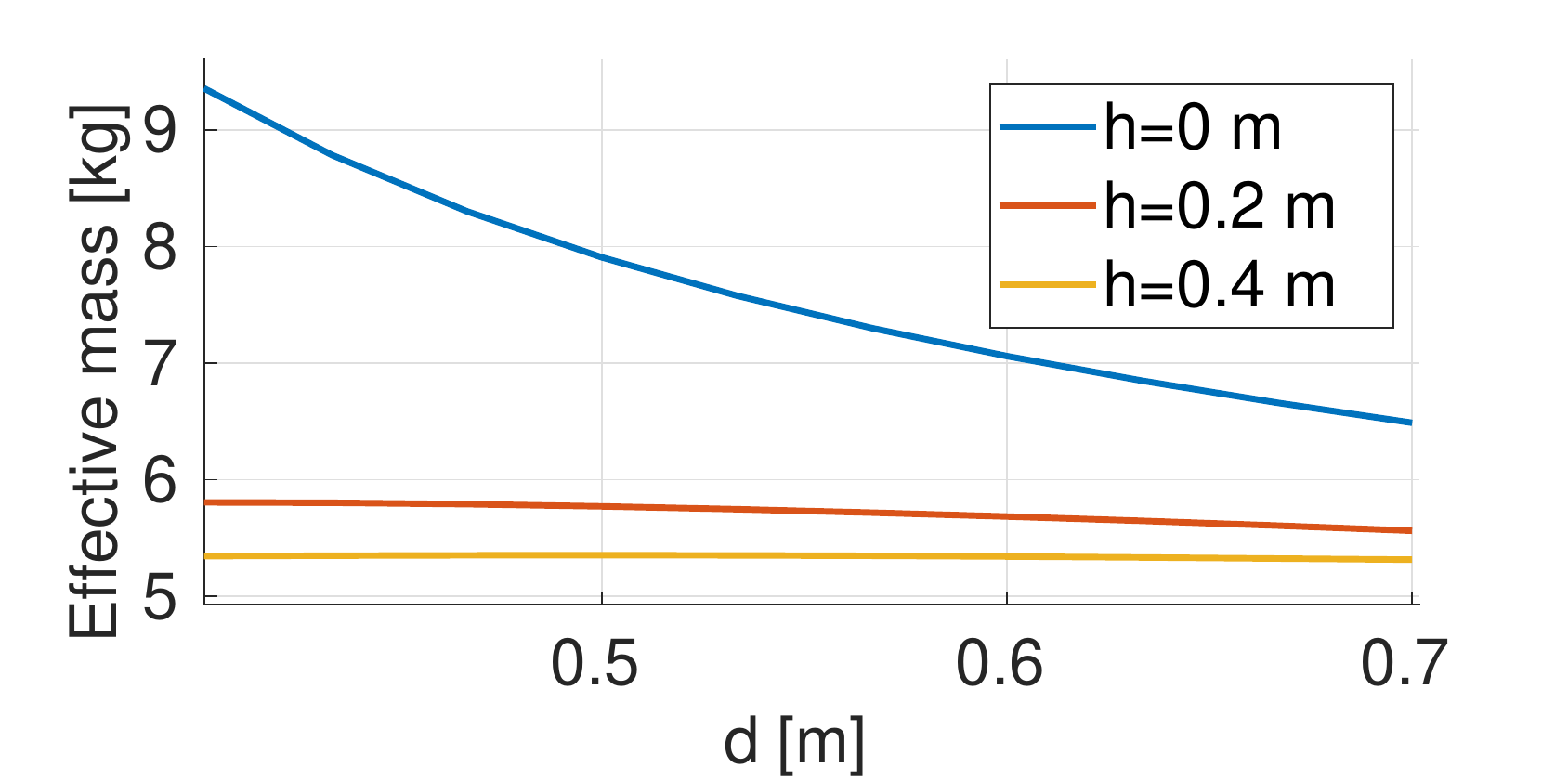}
\vspace{-0.6cm}
\caption{{\small $m_R$ as a function of $d$ and $h$.}}
\label{fig:3dof_height_effect}
\end{subfigure}
\vspace*{-5pt}
\caption{{\small Calculating effective mass of model 3 DoF planar manipulator. Collision direction ``down'': $u = [0,-1]$.}}
\label{fig:3dof}
\vspace{-0.6cm}
\end{figure}

Although the robots have 6 (UR10e) and 7 (KUKA) degrees of freedom (DoF), the robot configurations at impact can be coarsely approximated with a 3 DoF planar manipulator. Inspired by the UR10e manipulator, we used a model with three links with masses $[13, 4, 4]$ in $kg$ and the length of the links $[0.5, 0.45, 0.05]$ in $m$. On a grid resembling Fig.~\ref{fig:grid}, we used the analytical solution of inverse kinematics, restricted to the ``elbow up'' configuration, to reach with the EE the targets on the grid---see Fig.~\ref{fig:3dof_3by3}---and calculated the effective mass, with $u = [0,-1]$ (collision in the downward direction). We sampled the workspace more densely, giving rise to Fig.~\ref{fig:3dof_height_effect}, providing a prediction in line with Fig.~\ref{fig:force_distance_height}. The results also suggest that the effect of $d$ and $h$ should be considered together and ``cross-factors'' are needed. 

\subsection{Acquiring 3D Collision-Force-Map from data}
\label{subsec:acquire_3Dcfm}
We investigated the significance of every element of the model like the one in Eq.~\ref{eq:linear_model}, with additional terms in $h$ and terms with interaction factors between $d$, $h$, and $v$---for 3 datasets (Table~\ref{tab:dataset}) simultaneously using a two-stage process. We started with the polynomial model containing all terms (variables $d$, $v$, $h$) up to degree three and their interaction terms up to degree three (19 terms together). We removed all terms with a $p$-value higher than 0.05 for all three datasets in stage one, to obtain 13 terms for stage two. In stage two, we iteratively removed terms and compared two model parameters: Root Mean Squared Error ($RMSE$) and coefficient of determination ($R^2$). In every iteration, the fit would typically be worse and hence $RMSE$ would increase and $R^2$ decrease. The term for which its removal produced the smallest change of these two parameters was removed. The change was defined as follows: 
 $\sum_{\mathrm{datasets}}(\Delta RMSE + 100\Delta R^2)$.  The elimination procedure was stopped when this change for the term to be eliminated was bigger than 0.5 (i.e., removal of this term would make the fit significantly worse). The result of this process gave rise to Eq.~\ref{eq:our_model}:
\begin{eqnarray}
    \mathrm{ln}(F) =  \beta_{0} + \beta_{1} \cdot v + \beta_{2} \cdot d + \beta_{3}\cdot d^2 + \beta_{4}\cdot d\cdot h +  \nonumber \\ + \beta_{5}\cdot h^2 + \beta_{6}\cdot d^2 \cdot v + \beta_{7}\cdot d \cdot v^2 + \beta_{8}\cdot d \cdot h^2
    \label{eq:our_model}
\end{eqnarray}

\section{Experimental setup}
\label{sec:setup}

\subsection{Setup and robots}
\label{subsec:setup_and_robots}
An overview is in Fig.~\ref{fig:setup} and in the accompanying \href{https://youtu.be/4eHsbe4EuHU}{video}. The experiments consisted of a series of impacts with the robots at different locations in the workspace and different speeds onto an impact measuring device. Both robots were commanded using the Cartesian linear movement---where the EE follows a straight line---toward the impact. As a large number of impacts were performed (more than 400 per robot in total), we preferred not to use the robot flange but the surface at the last joint instead. 
Robots were controlled using their standard control interfaces while experimental data were collected. We also specify the safety settings used for the experiments as they influence the robots' overall behavior and, in particular, the response to a collision.

\paragraph{UR10e} Our UR10e is equipped with additional protective layer, Airskin, that is not used in this work but the extra weight (1.8 kg) is considered. The worst-case data collection frequency was 800 Hz. The second most restrictive safety setting was used, which restricts the robot mainly in force and speed, but allows for sufficient acceleration and deceleration with respect to our velocities.\footnote{Allowed power: 200 W, Momentum: 10 kg m/s, Stopping time: 300 ms, Stopping distance: 0.3 m, Tool speed: 0.75 m/s, Tool force: 120 N, Elbow speed: 0.75 m/s, Elbow force: 120N.}

\paragraph{KUKA LBR iiwa 7R800} The robot is equipped with a joint torque sensor in every axis and the safety setting restricts the maximal external torque at any joint. Two different settings were employed: 10 and 30 Nm. The data from the robot were collected with a frequency of 1000 Hz.

\subsection{Measuring device}
\label{subsec:measuring_device}
We used \textit{CBSF-75-Basic}---a measuring device (force transducer) for gauging forces and pressures, with a spring constant of 75000 N/m (see also \cite{Mewes2003}). Impact forces of up to 500 N can be measured, with maximum error up to 3 N (calibration protocol from supplier).
Following TS 15066, appropriate damping material was added to mimic impacts on the back of the non-dominant hand. 
Peak force from the impact Phase I (Fig.~\ref{fig:phases}) was recorded and used for analysis.

\subsection{Data collection}
\label{subsec:data_col}
Schlotzhauer et al.~\cite{Schlotzhauer2019} experimentally verified the rotational symmetry assumption. Thus, a single dimension, distance from the robot base, was the only relevant parameter. In our case, it is sufficient to study a plane in the 3D workspace, varying two dimensions: $d$ and $h$. 

\begin{figure}[tb]
    \centering
    \begin{subfigure}[t]{0.23\textwidth}
    \includegraphics[width=\textwidth]{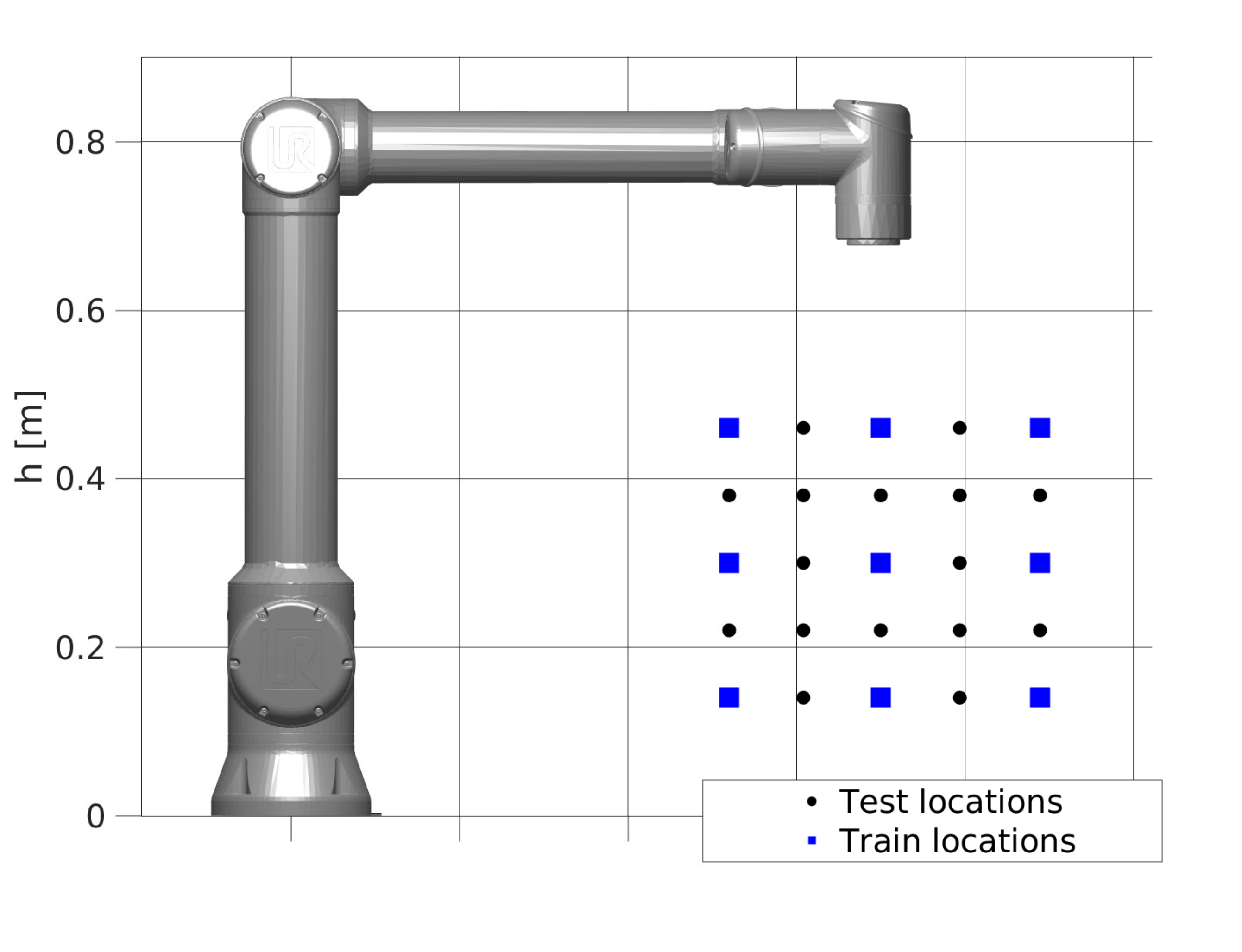} 
    \end{subfigure}
    ~
    \begin{subfigure}[t]{0.23\textwidth}
    \includegraphics[width=\textwidth]{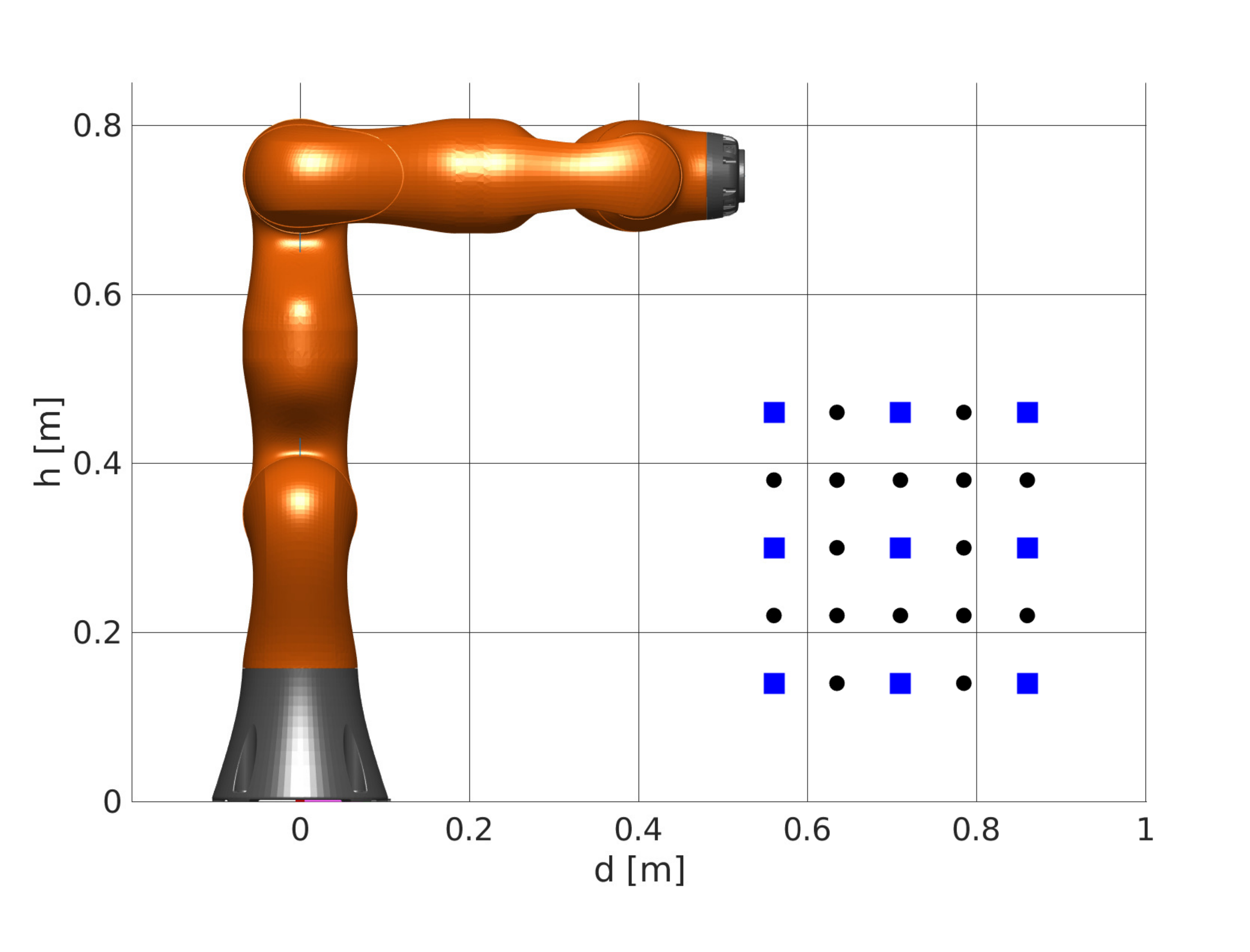} 
    \end{subfigure}
	\vspace*{-15pt}    
    \caption{{\small Measurements locations distribution. (Left) UR10e.\\ (Right) KUKA LBR iiwa 7R800.}}
    \label{fig:grid}
    \vspace{-0.4cm}
\end{figure}

For the UR robot, the $d$ ranged from 0.52 m to 0.89 m with increments of 0.09 m and five different heights from the level of the robot's base starting at 0.14 m with 0.08 m increments -- see Fig.~\ref{fig:grid} (Left). The KUKA robot has a different reach. We sampled the workspace at the following positions: $d$ from 0.56 m to 0.86 m with an increment of 0.075 m and five heights corresponding to heights used with UR robot -- see Fig.~\ref{fig:grid} (Right). At a given position, we performed measurements with five different velocities (0.20, 0.25, 0.30, 0.35, 0.40 m/s) in the downward direction. 

\paragraph{Training set} It is our goal to develop a practical tool that can be rapidly deployed. Therefore, the number of measurements needed should be as small as possible. For training the model, we use only a subset of the grid---9 locations with blue square markers in Fig.~\ref{fig:grid}---and 3 velocities (0.20, 0.30, 0.40 m/s).  
This gives rise to only 27 training measurements per robot, or 81 if every measurement is repeated 3 times. An overview is in Table~\ref{tab:dataset}.  
For the KUKA robot, the repeatability was higher. Hence, for the 10 Nm ext. torque setting, measurements were performed only once.

\begin{table}[]
\resizebox{240pt}{!}{%
\begin{tabular}{|c|c|c|c|}
\hline
\textbf{dataset} & \textbf{\begin{tabular}[c]{@{}c@{}}samples\\ per state\end{tabular}} & \textbf{\begin{tabular}[c]{@{}c@{}}training states\\ (used samples*)\end{tabular}} & \textbf{\begin{tabular}[c]{@{}c@{}}testing states\\ (used samples*)\end{tabular}} \\ \hline
\textbf{UR10e} & 3 & 27 (75) & 88 (249) \\ \hline
\textbf{KUKA 30 Nm} & 3 & 27 (78) & 98 (291) \\ \hline
\textbf{KUKA 10 Nm} & 1 & 27 (26) & 98 (98) \\ \hline
\end{tabular}
}
\caption{{\small Collected datasets. A ``state'' is a combination of distance, height, and speed. *Samples that exceeded the measuring device limit of 500 N were not used.}}
\label{tab:dataset}
\vspace{-0.4cm}
\end{table}


\paragraph{Testing set} For every robot, 16 additional positions (black dots in Fig.~\ref{fig:grid}) were tested with 5 velocities (0.20, 0.25, 0.30, 0.35, 0.40 m/s). For the 9 positions from the training set, only the velocities 0.25 and 0.35 m/s were added. In total, this gave 98 measurements per robot\footnote{Two positions (h = 0.3, d = 0.61; h = 0.3, d = 0.79) were omitted due to the experimenter's oversight on UR10e.}. 

\paragraph{Rotational symmetry verification} In order to verify the assumption that rotation of the first joint does not influence the results, 117 additional measurements on the UR10e robot were performed. 

All measurements above the recommended limit of the impact measuring device (500 N) were discarded.

\section{Experiments and results}
\label{sec:results}
Our results consist of a series of experiments in which two collaborative robots collide with an impact measuring device. An illustration is provided in the accompanying \href{https://youtu.be/4eHsbe4EuHU}{video}. First, we present and evaluate the 3D Collision-Force-Map model for the two robots. Second, we compare the results obtained with the 2D CFM \cite{Schlotzhauer2019} and the treatment of Power and Force Limiting in TS 15066 \cite{ISO/TS15066}. Finally, we present the force profiles after impact and analyze their implications.

\subsection{3D Collision-Force-Map for UR10e}
\label{subsec:ur_mes}

First, the rotational symmetry was experimentally verified using 117 measurements: 39 combinations of positions in the workspace and speed with 3 repetitions. The error, i.e. the difference in measured force on impact at different positions on the same circle (same $d$, $h$, and speed), was maximum 10 N (3.5 \%), mean 1 N (0.05 \%).



Second, restricting ourselves to a plane, we measured the impact forces on the grid of positions and at 5 different speeds (see Section~\ref{subsec:data_col}). Every measurement was repeated 3 times, with a maximum standard deviation (SD) of these three measurements of 3.85 N and a mean of these SDs across all locations/speeds of 1.12 N. 
In total, 324 measurements were performed. 

The training set was used to fit the model of the form in Eq. \ref{eq:our_model}. The obtained model was:
\begin{eqnarray}
\mathrm{ln}(F) = 6.2990 + 3.3761\cdot v - 1.1050\cdot d - \nonumber \\ - 1.3066 \cdot d^2  - 1.5258 \cdot d \cdot h - 6.6954 \cdot h^2 + \\ + 4.0919 \cdot d^2 \cdot v - 6.0090 \cdot d \cdot v^2 + 8.5207 \cdot d \cdot h^2 \nonumber
    \label{eq:ur_model}
\end{eqnarray}

\begin{figure}[tb]
	\centering
	\includegraphics[width=0.32\textwidth]{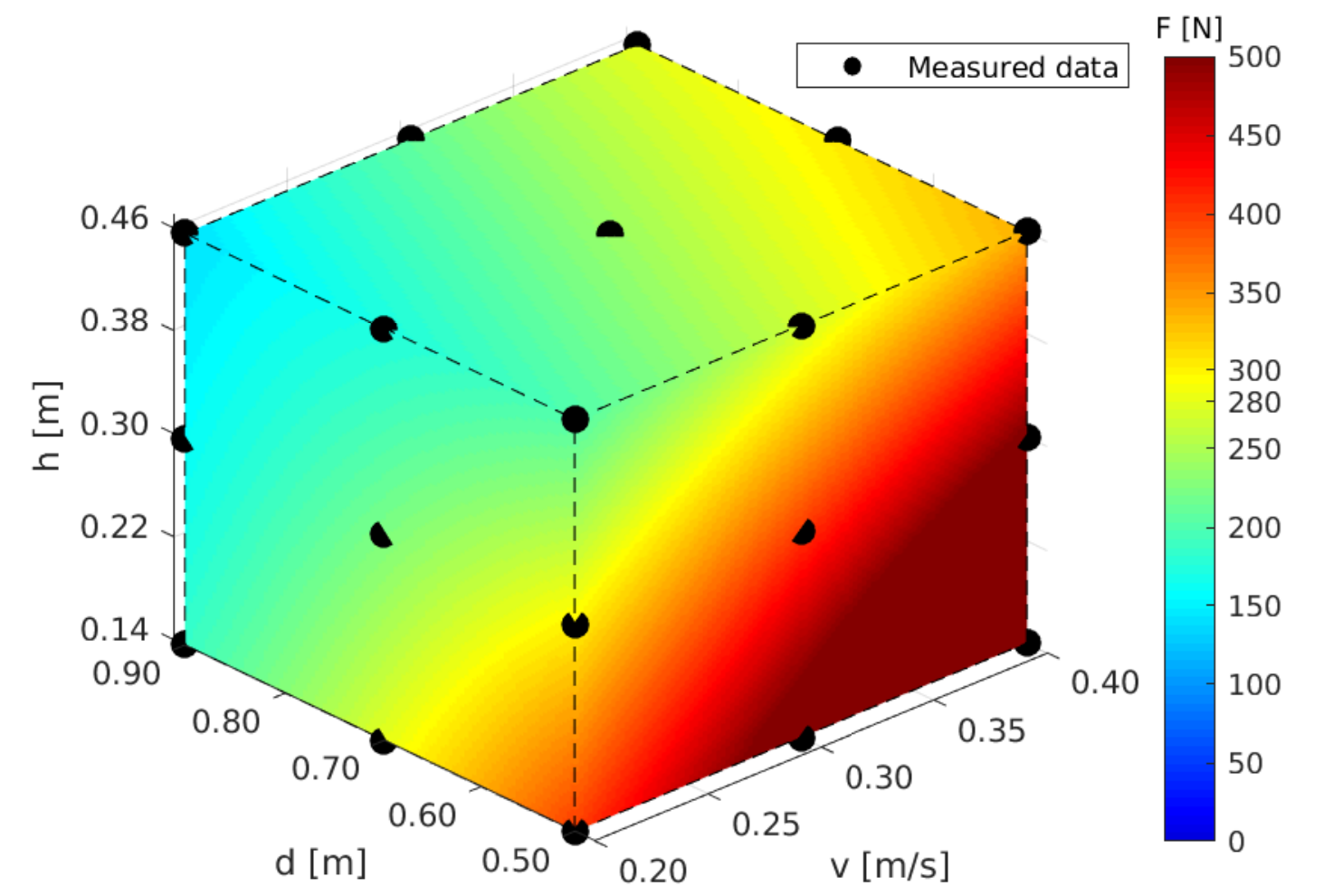}
	\caption{{\small UR10e -- 4D visualization of 3D CFM model predictions within the sampled robot workspace for different collision velocities. The robot base is located at $d=0$ and $h=0$. Black points are showing measured training data capturing the measurement grid from Fig. \ref{fig:grid} (Left).}}
	\label{fig:ur_cube}
	\vspace{-0.4cm}
\end{figure}

Figure~\ref{fig:ur_cube} shows the three variables, $h$, $d$, and $v$; only the surface of this color map is visible though. As would be expected, impact forces are directly proportional to the velocity. For a fixed $v$ and $h$ or $d$, a 2D visualization is possible---green lines in Fig.~\ref{fig:cfm_comp}.

Table \ref{tab:results} quantifies the accuracy of our model on the testing set (we refer to the UR10e row here). We evaluate underestimation and overestimation of the impact forces separately---the former being more critical regarding safety assessment of the application. The maximal underestimation over the testing set (see Table~\ref{tab:dataset}) is 3.56 \% (8.42 N) and the mean underestimation is 1.37 \% (4.45 N). Overestimation is higher, 6.35 \% (22.45 N) at maximum and 2.40 \% (6.97 N) on average. 
The error over the whole 3D CFM dataset is underestimating slightly more, 3.68 \% (7.07 N). Overestimation is more frequent for higher force values and underestimation for lower impact forces. The higher relative overestimation is probably due to the lower density of measurements for higher forces---when impact forces surpassed 500 N.

With a bound on the underestimation, the 3D CFM can be used to determine a safe speed for a collaborative application. Adding 10\% to all predicted forces---a conservative choice---and knowing $d$, $h$, and allowed impact forces, one can rearrange Eq.~\ref{eq:our_model} to obtain the maximum safe EE velocity.  

\begin{table}[bt]
\resizebox{240pt}{!}{%
\begin{tabular}{|c c|c|c|c|c|}
\hline
\multicolumn{2}{|c|}{\textbf{dataset}}  & \textbf{\begin{tabular}[c]{@{}c@{}}max UE\\  {[}\% / N{]}\end{tabular}} & \textbf{\begin{tabular}[c]{@{}c@{}}mean UE \\ {[}\% / N{]}\end{tabular}} & \textbf{\begin{tabular}[c]{@{}c@{}}max OE\\ {[}\% / N{]}\end{tabular}} & \textbf{\begin{tabular}[c]{@{}c@{}}mean OE\\  {[}\% / N{]}\end{tabular}} \\ \hline
\multirow{2}{*}{\textbf{UR10e}} & \textbf{Ts} & 3.56 / 8.42 & 1.37 / 4.45 & 6.35 / 22.45 & 2.40 / 6.97 \\ \cline{3-6} 
 & \textbf{All} & 3.68 / 7.07 & 1.50 / 4.63 & 6.35 / 22.45 & 2.16 / 6.26 \\ \hline
\multirow{2}{*}{\textbf{Kuka 30 Nm}} & \textbf{Ts} & 9.30 / 22.32 & 2.63 / 7.61 & 9.40 / 16.54 & 3.12 / 8.07 \\ \cline{3-6} 
 & \textbf{All} & 9.30 / 22.32 & 2.58 / 7.37 & 9.40 / 16.54 & 3.08 / 7.93 \\ \hline
\multirow{2}{*}{\textbf{Kuka 10 Nm}} & \textbf{Ts} & 5.76 / 19.36 & 1.96 / 5.18 & 5.02 / 14.71 & 1.63 / 4.16 \\ \cline{3-6} 
 & \textbf{All} & 5.93 / 20.20 & 1.94 / 5.26 & 5.38 / 19.48 & 1.59 / 4.11 \\ \hline
\end{tabular}%
}
\vspace{-0.1cm}
\caption{{\small Accuracy of 3D CFM model with underestimation (\textbf{UE}), overestimation (\textbf{OE}), the test set (\textbf{Ts}), complete dataset (\textbf{All}).}}
\label{tab:results}
\vspace{-0.2cm}
\end{table}

\subsection{3D Collision-Force-Map for KUKA LBR iiwa 7R800}
\label{subsec:meas_KUKA}
Similarly to the UR robot, restricted to a plane, we measured impact forces on the grid of positions and at 5 different speeds (Section~\ref{subsec:data_col}, Fig.~\ref{fig:grid} (Right), Table~\ref{tab:dataset} for details). We collected measurements with two different safety settings (30 and 10 Nm of max. external torque at any joint).
\paragraph{30 Nm external torque setting} Every measurement, same location and speed, was repeated 3 times, with maximum SD of 3.09 N. The mean of these SDs across all locations/speeds was 0.58 N. The dataset is composed of 369 measurements (see Table~\ref{tab:dataset}). The model, 3D CFM, for this robot and settings is given by the equation:
\begin{eqnarray}
\mathrm{ln}(F) = 7.0641 + 4.2943\cdot v - 4.5286\cdot d + \nonumber \\ + 0.9917 \cdot d^2 - 0.5795 \cdot d \cdot h - 6.0074 \cdot h^2 + \\ + 3.9366 \cdot d^2 \cdot v - 7.2169 \cdot d \cdot v^2 + 7.0446 \cdot d \cdot h^2 \nonumber
    \label{eq:kuka30_model}
\end{eqnarray}

\begin{figure}[tb]
	\centering
	\includegraphics[width=0.48\textwidth]{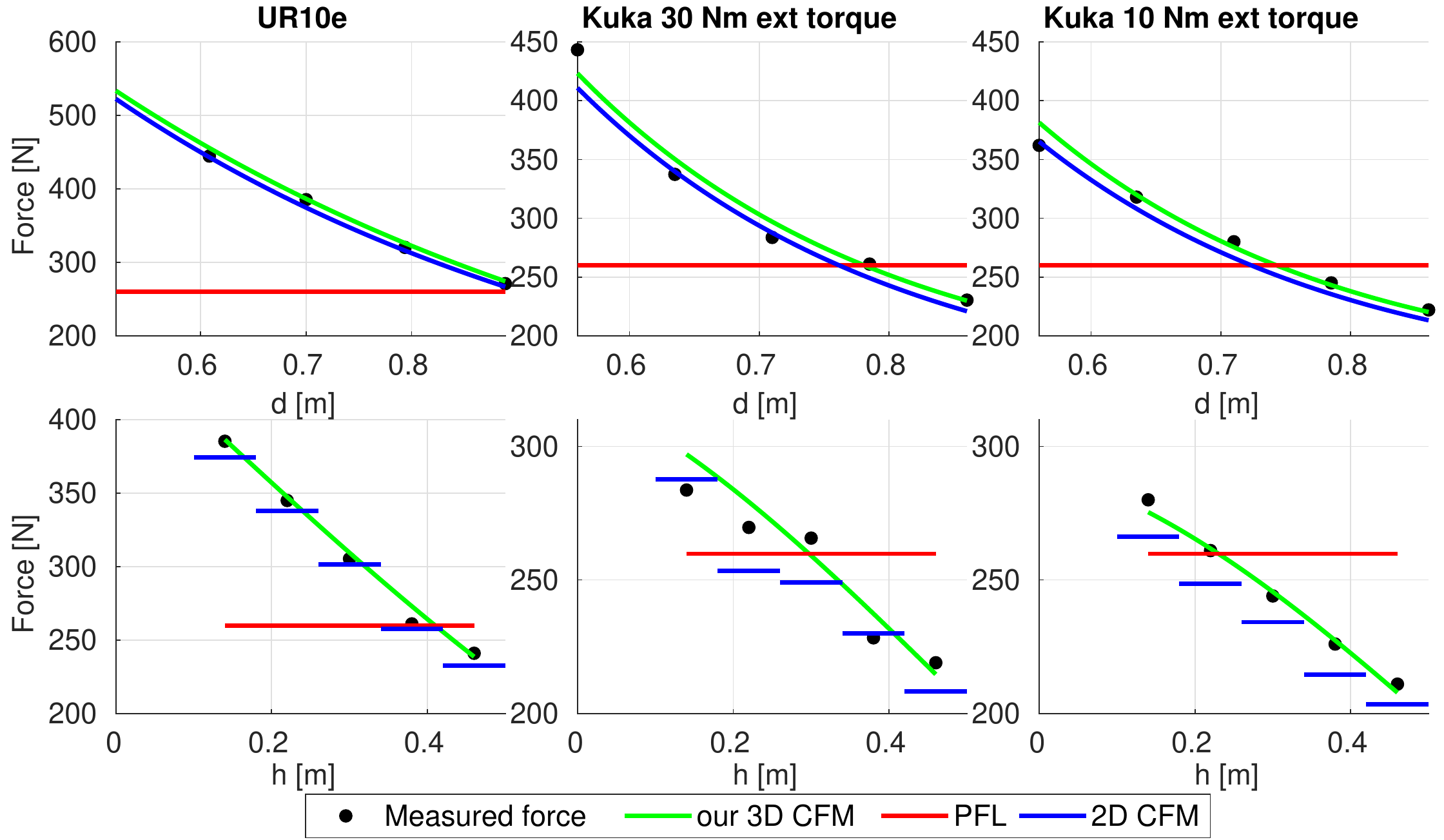}
	\vspace*{-10pt}
	\caption{{\small Impact force model comparison of 3D CFM (ours) -- green, 2D CFM \cite{Schlotzhauer2019} -- blue, and the value for Power and Force Limiting mode from \cite{ISO/TS15066} -- red. EE velocity = 0.30 m/s. (Top) Constant height of end effector in the workspace (0.14 m). (Bottom) Constant distance from base (0.70 m for UR10, 0.71 m for KUKA).}}
	\label{fig:cfm_comp}
	\vspace{-0.6cm}
\end{figure}

The results for one speed (0.30 m/s) are shown in Fig.~\ref{fig:cfm_comp}, center, with the green line---with fixed height (0.14 m, top) or distance (0.71 m, bottom). 

The accuracy of the model is quantified in Table~\ref{tab:results}. The maximal underestimation error is 9.30 \% (22.32 N) with a mean underestimation of 2.63 \% (7.61 N) over the testing set and 2.58 \% (7.37 N) over the whole 3D CFM dataset. The overestimation is comparable to the underestimation, with a mean value of 3.12 \% (8.07 N) over the testing set and 3.08 \% (7.93 N) over the whole 3D CFM dataset, and with the maximal error of 9.40 \% (16.54 N). Both under- and overestimation are worse than in the case of the UR robot. 

\paragraph{10 Nm external torque setting} Due to the high repeatability of the measurements in the 30 Nm setting, deviations under the precision of the measurement device, only one measurement per position and speed was taken. 
The resulting dataset contains 124 measurements (see Table~ \ref{tab:dataset}). 
The resulting model, 3D CFM, for KUKA with 10 Nm is: 
\begin{eqnarray}
\mathrm{ln}(F) = 6.6936 + 4.9297\cdot v - 4.4782\cdot d + \nonumber \\ + 1.2926 \cdot d^2 - 0.3758 \cdot d \cdot h - 5.5669 \cdot h^2 + \\ + 3.2609 \cdot d^2 \cdot v - 7.2332 \cdot d \cdot v^2 + 6.4016 \cdot d \cdot h^2 \nonumber.
    \label{eq:kuka10_model}
\end{eqnarray}
The results for one speed (0.30 m/s) are shown in Fig.~\ref{fig:cfm_comp}, right, with the green line---with fixed height (0.14 m, top) or distance (0.71 m, bottom). Compared to the 30 Nm setting, the forces are on average approximately 5\% lower. 
The accuracy of the model is quantified in Table~\ref{tab:results}. The maximal underestimation is lower than with the previous safety settings with 5.76 \% (19.36 N) and an average of 1.96 \% (5.18 N) over the testing set and 1.94 \% (5.26 N) over the whole 3D CFM dataset. The overestimation is even lower than with the UR robot with a peak value of 5.02 \% (14.71 N) and 1.63 \% (4.16 N) on average over the testing set and 1.59 \% (4.11 N) over the whole 3D CFM dataset. 

\subsection{3D Collision-Force Map vs. 2D CFM vs. PFL (TS 15066)}
First, we want to compare our results with the 2D Collision-Force-Map (2D CFM) \cite{Schlotzhauer2019}. We used the least-squares method to train the 2D CFM model (Eq. \ref{eq:linear_model}) on our data, using the 0.20, 0.30, and 0.40 m/s EE velocities. 
A comparison for one velocity (0.30 m/s) and one height (0.14 m) is visualized in the top panels of Fig.~\ref{fig:cfm_comp}. As the 2D CFM model does take $h$ into account and as we have shown the forces to importantly depend on this parameter, a single 2D CFM model will fail to deliver predictions on the whole workspace. To allow for a more fair comparison, we have trained it separately for every height---blue lines in the bottom panels of Fig.~\ref{fig:cfm_comp}.  
As can be seen in Table \ref{tab:results_2d}, the 2D CFM overestimation errors are comparable to our 3D CFM model errors (higher for UR and lower for KUKA 30 Nm dataset). On the other hand, the 2D CFM models underestimate significantly more than our 3D CFM model. 


\begin{table}[bt]
\resizebox{240pt}{!}{%
\begin{tabular}{|c c|c|c|c|c|}
\hline
\multicolumn{2}{|c|}{\textbf{dataset}}  & \textbf{\begin{tabular}[c]{@{}c@{}}max UE\\  {[}\% / N{]}\end{tabular}} & \textbf{\begin{tabular}[c]{@{}c@{}}mean UE \\ {[}\% / N{]}\end{tabular}} & \textbf{\begin{tabular}[c]{@{}c@{}}max OE\\ {[}\% / N{]}\end{tabular}} & \textbf{\begin{tabular}[c]{@{}c@{}}mean OE\\  {[}\% / N{]}\end{tabular}} \\ \hline
\multirow{2}{*}{\textbf{UR10e}} & \textbf{Ts} &\cellcolor{gray!30} 6.51 / 14.60 & \cellcolor{gray!30}2.23 / 6.33 &\cellcolor{gray!30} 7.56 / 11.22 & \cellcolor{gray!30}1.76 / 5.17\\ \cline{3-6} 
 & \textbf{All} & \cellcolor{gray!30} 6.51 / 14.60 & \cellcolor{gray!30}2.27 / 6.42 & \cellcolor{gray!30}9.13 / 13.21 &\cellcolor{gray!30} 2.78 / 7.98 \\ \hline
\multirow{2}{*}{\textbf{KUKA 30 Nm}} & \textbf{Ts} & \cellcolor{gray!30}12.01 / 28.83& \cellcolor{gray!30}3.94 / 10.91 & 7.32 / 27.67 & 2.73 / 7.19\\ \cline{3-6} 
 & \textbf{All} & \cellcolor{gray!30} 12.01 / 28.83 & \cellcolor{gray!30} 3.75 / 10.37 & 7.89 / 22.14 & 2.84 / 7.46 \\ \hline
\multirow{2}{*}{\textbf{KUKA 10 Nm}} & \textbf{Ts} & \cellcolor{gray!30} 8.70 / 29.23 & \cellcolor{gray!30} 3.04 / 7.59 & \cellcolor{gray!30} 5.24 / 21.76 & \cellcolor{gray!30}2.18 / 5.82 \\ \cline{3-6} 
 & \textbf{All} & \cellcolor{gray!30} 8.70 / 29.23 & \cellcolor{gray!30} 3.00 / 7.53 & 5.24 / 21.76 & \cellcolor{gray!30} 2.12 / 5.43 \\ \hline
\end{tabular}%
}
\caption{{\small Accuracy of 2D CFM models. Gray values indicate worse performance than 3D CFM model (Table~\ref{tab:results}).}}
\label{tab:results_2d}
\vspace{-0.6cm}
\end{table}

Power and Force Limiting according to \cite{ISO/TS15066} does not take $d$ or $h$ into account and considers velocity only (see Section~\ref{subsec:PFL}). Eq.~\ref{eq:v_pfl} can be rearranged and $F$ obtained. With the corresponding robot masses and $v=0.3$, this gives rise to the red lines in Fig.~\ref{fig:cfm_comp}. 
Clearly, such an approximation is insufficient. Moreover, next to overestimation, it leads also to gross underestimation of the impact forces and hence violates the safety of the human (by the very standards of \cite{ISO/TS15066}). 

\subsection{Nature of dynamic impact}
\label{subsec:clamping}

\begin{figure}[tb]
	\centering
	\includegraphics[width=0.4\textwidth]{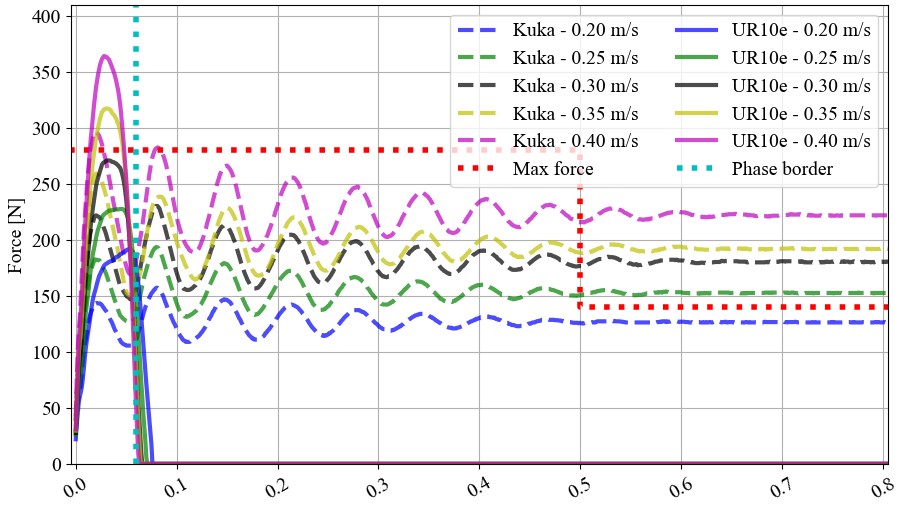}
	\vspace*{-7pt}
	\caption{{\small Force evolution after impact. UR10e (at $d=0.89\ \mathrm{m}$, $h=0.14\ \mathrm{m}$) -- solid lines. KUKA (at $d=0.86\ \mathrm{m}$, $h=0.14\ \mathrm{m}$, $30\ \mathrm{Nm}$ ext. torque limit) -- dashed lines. Phase I / Phase impact phase boundary (cf. Fig~\ref{fig:phases}) -- cyan dotted line. Permissible force per TS 15066 -- red dotted line.}}
	\label{fig:robots_force_comp}
	\vspace{-0.7cm}
\end{figure}

Peak force estimation is only one component required to assess safety of a HRC application. Collision force evolution after ``Phase I'' (Section~\ref{subsec:PFL}, Fig.~\ref{fig:phases}) is also important. Fig.~\ref{fig:robots_force_comp} shows this for a selection of our experiments. For the UR10e robot, only Phase I is present. That is, although the scenario has a ``clamping nature'', the UR10e controller makes the EE actively bounce back and thus makes the actual contact of a transient kind. On the other hand, the KUKA robot shows a prolonged damped harmonic movement upon impact.


TS 15066 prescribes maximum force thresholds for the first 0.5 s of impact (transient contact) and half this threshold afterward (quasi-static contact)---as shown in Fig.~\ref{fig:robots_force_comp} with red dotted lines. Thus, based on our empirical findings, one could apply the higher force thresholds for the UR10e (e.g., 280 N) and only half that threshold for the KUKA (140 N), which would dramatically alter the safe speeds in the application.

\section{Conclusion, Discussion, Future Work}
Using two collaborative robots, UR10e and KUKA LBR iiwa 7R800, we performed 934 measurements of forces exerted on the impact of the robot end-effector with an impact measuring device, with different robot velocities and at different locations in the robot workspace. The collision direction was always down, perpendicular to the table surface. We established a clear relationship between the distance from the robot base and the impact forces (in line with \cite{Schlotzhauer2019}) and, newly, also the height in the workspace---both variables being inversely proportional to the impact forces. We developed a data-driven model---3D Collision-Force-Map---that estimates the forces as a function of distance, height, and velocity, including their mutual relationships. This model is more accurate than 2D CFM \cite{Schlotzhauer2019} and PFL according to \cite{ISO/TS15066} that does not take position in the workspace into account. Furthermore, we show that it can be trained from a limited amount of data: we sampled only 9 positions in the workspace and 3 velocities to train the model.

Thus, our main contribution is a tool that allows for rapid prototyping of a collaborative robot workspace. For quasi-static impacts on the back of the hand, a force limit of 140 N is prescribed by TS 15066~\cite{ISO/TS15066}, which would based on the formula from TS 15066 limit the allowed EE speed in the whole workspace to 0.13 and 0.16 m/s for the UR10e and KUKA LBR iiwa, respectively. Our measurements reveal that if the task is performed, for example, 0.8 m away and 0.4 m above the robot base, speeds of 0.16 m/s (UR10e) and 0.20 m/s (KUKA LBR iiwa) stay within the prescribed force limit. Furthermore, we observe that despite the clamping nature of our scenario, the UR10e robot generates only transient contact. With the 280 N limit, 0.36 m/s will still be safe with the UR10e---an almost threefold increase. The PFL formulas from TS 15066 are insufficient---leading both to significant underestimation and overestimation at different locations in the robot workspace, and thus to unnecessarily long cycle times or even dangerous applications. 
The impact measuring device was firmly attached to the table. The possibility that a human operator would be moving against the robot prior to collision was thus not considered. However, we focused on quasi-static contacts where the limits are  stricter. The most dangerous part of the incident is in the clamping nature.

Interestingly, the trend of the relationship between distance from / height above the robot base and the forces exerted on collision is largely consistent across two different collaborative robots and also in line with our simple 3 DoF model. However, whether the effective mass entirely determines the trends in empirically measured forces remains an open question. In addition, the impacts are not uncontrolled. The collision is detected by the robot internal controllers, generating a response, which is likely quick enough to shape the force evolution even during the first phase of the impact. Extending the effective mass models is thus impeded by the fact that accurate inertial parameters of the manipulators are not known and the controllers are proprietary. In our case, different safety settings (external torque) resulted in different impact forces. Thus, empirical assessment of impact forces in the robot workspace seems indispensable at the moment. 

It should be noted that our results are not expected to generalize to other robots or even different collision sites,  directions, or kinematic configurations on the same manipulators. We concentrated on downward movement of the robot to the table, which is typical of many applications, and quasi-static contact, which represents the worst-case scenario. Impacts were made with the last robot joint, not the flange, for practical reasons. We propose an empirical method that can be deployed by robot integrators on a specific application site to quickly determine the optimal speed and position in the workspace where a task can be safely performed with maximum efficiency. The contact type and location on the robot and position in the workspace should all be set according to the application. In summary, for effective design of a collaborative application, empirical measurements are indispensable.

\bibliographystyle{IEEEtran}
\bibliography{3DcollisionForceMap}

\end{document}